\documentclass{article}

\usepackage{arxiv}

\usepackage[utf8]{inputenc} 
\usepackage[T1]{fontenc}    
\usepackage[colorlinks=false]{hyperref}       
\usepackage{url}            
\usepackage{booktabs}       
\usepackage{amsfonts}       
\usepackage{nicefrac}       
\usepackage{microtype}      
\usepackage{lipsum}		
\usepackage{graphicx}
\usepackage{doi}
\usepackage[style=apa]{biblatex}
\usepackage{placeins}

\title{Developing synthetic microdata through machine learning for firm-level business surveys}

\author{ 
\hspace{1mm}Jorge~Cisneros
\\
	Oak Ridge Institute for Science \& Education, USA\\
	\And
	\href{https://orcid.org/0000-0001-9863-1457}{\includegraphics[scale=0.06]{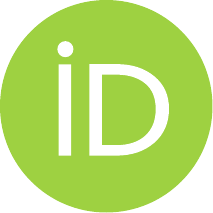}\hspace{1mm}Timothy~Wojan} \thanks{Corresponding Author \texttt{tim.steenburgh@gmail.com}} \\
	Oak Ridge Institute for Science \& Education, USA \\
	\And
    \href{https://orcid.org/0000-0001-8894-1240}{\includegraphics[scale=0.06]{orcid.pdf}\hspace{1mm}Matthew~Williams} \\
	RTI International, USA\\
    \And
    \hspace{1mm}Jennifer~Ozawa
    \\
	RTI International, USA\\
    \And
    \href{https://orcid.org/0000-0002-6979-1766}{\includegraphics[scale=0.06]{orcid.pdf}\hspace{1mm}Robert~Chew} \\
	RTI International, USA\\
    \And
    \href{https://orcid.org/0009-0004-5535-4620}{\includegraphics[scale=0.06]{orcid.pdf}\hspace{1mm}Kimberly~Janda} \\
	RTI International, USA\\
    \And
    \hspace{1mm}Timothy~Navarro
    \\
	RTI International, USA\\
    \And
    \hspace{1mm}Michael~Floyd
    \\
	Knexus Research Corporation, USA\\
    \And
    \hspace{1mm}Christine~Task
    \\
	Knexus Research Corporation, USA\\
    \And
    \hspace{1mm}Damon~Streat
    \\
	Knexus Research Corporation, USA\\
}

\renewcommand{\shorttitle}{Synthetic Business Microdata}

\begin{document}
\maketitle

\begin{abstract}
	Public-use microdata samples (PUMS) from the United States (US) Census Bureau on individuals have been available for decades. However, large increases in computing power and the greater availability of Big Data have dramatically increased the probability of re-identifying anonymized data, potentially violating the pledge of confidentiality given to survey respondents. Data science tools can be used to produce synthetic data that preserve critical moments of the empirical data but do not contain the records of any existing individual respondent or business. Developing public-use firm data from surveys presents unique challenges different from demographic data, because there is a lack of anonymity and certain industries can be easily identified in each geographic area. This paper briefly describes a machine learning model used to construct a synthetic PUMS based on the Annual Business Survey (ABS) and discusses various quality metrics. Although the ABS PUMS is currently being refined and results are confidential, we present two synthetic PUMS developed for the 2007 Survey of Business Owners, similar to the ABS business data. Econometric replication of a high impact analysis published in \emph{Small Business Economics} demonstrates the verisimilitude of the synthetic data to the true data and motivates discussion of possible ABS use cases.
\end{abstract}

\setcounter{footnote}{0}

\section{Introduction}
In partnership between the US Census Bureau and the NCSES, the Annual Business Survey (ABS) captures essential firm details and key metrics that provide insights into the diverse economic landscape of the United States \footnote{\href{https://www.census.gov/programs-surveys/abs.html}{census.gov/programs-surveys/abs.html}}\textsuperscript{,} \footnote{\href{https://ncses.nsf.gov/surveys/annual-business-survey/2022}{ncses.nsf.gov/surveys/annual-business-survey/2022}}. This mandatory survey collects data on innovation, globalization, and business owner characteristics for nonfarm, for-profit businesses and measures R\&D expenditures for a subset of those businesses with 1 to 9 employees. The ABS frame includes about 5 million businesses engaged in a variety of industries including, but not limited to mining, utilities, construction, manufacturing, wholesale trade, retail trade, and services industries. The ABS sample consists of approximately 850,000 firms every 5 years and approximately 300,000 firms every year in between. The sample itself includes all companies classified in selected research-intensive industries, such as scientific R\&D services. The data collected through the ABS is crucial for policymakers, economists, businesses, and academic researchers to analyse trends, assess business performance, and develop strategies for economic growth and development \parencite{autor_fall_2020, decker_role_2014, haltiwanger_high_2016, mitchell_track_2017, zolas_advanced_2020}. 

The US Foundations for Evidence-Based Policymaking Act of 2018, also known as the Evidence Act, requires federal statistical agencies in the United States to make data more accessible to the public and available in multiple access tiers based on data sensitivity \parencite{rep_ryan_hr4174_2019}. The US CHIPS and Science Act, signed into law in 2022, authorized the NSF to establish the National Secure Data Service Demonstration Project (NSDS-D) to further the Evidence Act goals of accessibility with historic investments in curiosity-driven research and use-inspired, translational research \parencite{rep_ryan_hr4346_2022}. The NSDS-D will provide a platform to streamline and innovate data access and privacy protection to support expanded data use and inform research and policy discussions. Currently, the only public-use data files for the ABS are through tabulated data, while the richer ABS microdata is restricted-use and must adhere to confidentiality provisions under US Code Title 13 and US Code Title 26. This microdata file is accessible in-person within one of the 34 Federal Statistical Research Data Centers (FSRDCs) scattered across the country with an approved request through the standard application process (SAP) that US statistical agencies have coordinated for accessing confidential datasets.  Remote access to these datasets in the FSRDCs is currently being evaluated in a pilot program. Applicants must obtain Special Sworn Status (SSS) through in-depth background investigations with processing times of 3 to 4 months, must have lived in the US for 36 of the past 60 months, and must have a US-based institutional affiliation. Foreign national applications are welcome but require additional processing and take longer to complete. Moreover, successful applicants must take annual data stewardship training and are prohibited from discussing or sharing results with other colleagues without SSS until cleared by the FSRDC Disclosure Avoidance Officer or US Census Bureau Disclosure Review Board.

Public-use microdata samples (PUMS) therefore play a critical role in meeting the demand for public access to government-funded data and are essential for exploratory research and decision-making at vital academic, government, and industry levels \parencite{foster_panel_2020, hitt_health_2016, nick_hart_barriers_2018}. A PUMS can enable statistical organizations to reach a wider public and support more use cases in a timely manner with data products, decreasing the reliance on FSRDCs for high-level analyses and increasing public access to vital data \parencite{buffington_stem_2016, foster_innovation_2019, goldschlag_research_2019, jin_economies_2017, zolas_wrapping_2015}. The United Nations Economic Commission for Europe (UNECE) emphasizes that a PUMS is helpful for testing code before submitting a slow-moving request to run on internal data and testing analyses before deciding which procedures to submit to a validation server in a secure environment, all while giving rise to transparency \parencite{christopher_jones_synthetic_2022, gjaltema_high-level_2022}. Even so, there are limited PUMS for business surveys, either at the firm or establishment-level \parencite{seamans_ai_2018}. The Survey of Business Owners (SBO) collected data on economic and demographic characteristics for businesses and their owners in the US but was discontinued after the 2012 survey year \footnote{\href{https://www.census.gov/programs-surveys/sbo.html}{https://www.census.gov/programs-surveys/sbo.html}}. This data is now collected as part of the ABS. That same year, the first and only SBO PUMS for 2007 was released with disclosure techniques, such as data suppression, data modification, and noise infusion, that are not private enough to continue applying for modern standards and attacks \parencite{abowd_modernization_2020, doyle_confidentiality_2001, lauger_disclosure_2014, us_census_bureau_2007_2007}. The SBO 2007 PUMS is still requested today, indicating healthy demand for 18-year-old firm-level microdata and a potentially large latent demand for an update.  

The open challenge remains releasing an updated business PUMS that preserves privacy and is analytically useful \parencite{decker_declining_2017, haltiwanger_who_2013}. In the last decade, artificial intelligence (AI) has exploded in nearly every sector of industry, ranging from marketing to agriculture to biomedical imaging. Given its power to effectively learn data distributions, a natural next step to tackle our open challenge is to generate artificial data that not only represents the statistical properties of restricted-use data, but also intrinsically ensures privacy and confidentiality of sensitive information. 

\section{PUMS for Business Survey Data}
De-identified data, particularly in the context of PUMS, refers to anonymized data that has been processed with the intention of preventing re-identification of individual records in the dataset. Effective de-identification approaches provide both good privacy and good utility, such that a useful PUMS is one that is successful against individual re-identification, while preserving the moments and cross-product moments of the original restricted-use data. Developing a PUMS for business survey data presents unique challenges compared to demographic survey data, since the scale and variability across businesses are typically more diverse and varied than people and there is already a substantial amount of public data on businesses \parencite{bartelsman_microdata_2023, vilhuber_synthetic_2016}. The size and number of establishments of firms vary widely, with a small percentage of large firms often accounting for a significant portion of employment, sales, and other metrics. This skewness in the data distribution requires careful handling to ensure that the resulting PUMS maintains the statistical integrity of the restricted-use data \parencite{crane_business_2019}. 

Moreover, business survey data contains numerous industry-specific and geographic features that add complexity to the de-identification process, and, therefore, utility of the PUMS. Releasing the combination of fine-detail industry information, e.g. 6-digit NAICS (North American Industrial Classification System) codes, with fine-detail geographic data, e.g. zip codes, yields an incredibly useful PUMS, but highly revealing of individual firms. On the opposite extreme, a PUMS with highly aggregated industry and geographic information preserves the privacy of the firms but is of limited use. Re-identification risks further increase as the amount of data grows and becomes more widely available. Privacy can also be breached when numerous features of a firm within the PUMS are linked with other publicly available data, such as the Securities \& Exchange Commission (SEC) filings, County Business Patterns, and National Establishment Time Series. There is a clear demand for thorough and rigorous testing to assess re-identification risks.

\section{De-identification}
Common de-identification methods fall short for business survey microdata such as that collected in the ABS. The more information added with standard approaches, the easier it becomes to entirely omit or identify firms via other public or proprietary data. As a simple yet general example, suppose a retail store owned by a Hispanic female of age 52 in Austin, TX specializes in sustainable clothing, has existed for the last 20 years, and has an annual revenue of \$150,000. If the ABS PUMS only has a handful of features like owner sex, 2-digit NAICS code, and MSA (metropolitan statistical area), it is easy to “hide” this firm among the rest from the US via common de-identification techniques like cell suppression, subsampling, or even differential privacy: Austin is a large metropolitan with numerous female-owned retail businesses. However, if we begin to include other important features like owner race, age of the business, annual revenue, and number of employees, or provide finer details like 4-digit NAICS codes instead of 2-digit, then de-identification methods fail and will either omit and give empty responses to those features (cell suppression), imply there are zero firms with nonzero responses (subsampling), or give substantially different responses (differential privacy). See \textcite{das2024, vilhuber_synthetic_2016, hu_advancing_2024} for comparisons of these methods. 

Collecting complete feature information for every firm rapidly becomes difficult to provide in a way that ensures good and equitable utility for all users. Indeed, we see this phenomenon with demographic microdata as well, but business microdata has substantial re-identification risk due to public media data (newspaper articles, award ceremonies, marketing, \emph{etc}.) and public-use government-released data (longitudinal data, 4 or 6-digit NAICS codes from other mandatory surveys, \emph{etc}.). When variables or responses are removed from the PUMS to protect businesses’ privacy, the PUMS can lose its utility and make it impossible to conduct downstream analyses. In order to increase utility of the PUMS, for each  record, we want to capture correlations between features for their subpopulation, so we can maintain the original data distribution while safeguarding against re-identification risks inherent in detailed business microdata. This challenge underscores the need for innovative solutions, such as the use of synthetic data, to balance the demands for rich analytical insights with robust privacy protection measures without requiring hyper-specific frameworks that do not generalize or are non-intuitive.

\section{Synthetic Data and AI}
In simple terms, synthetic data is artificially generated data that represents statistical properties of real data, while ensuring privacy and confidentiality of sensitive information. Maintaining the statistical relationships of the original data enables synthetic data’s high utility and meaningful analyses. Unlike modified restricted-use data, synthetic data preserves privacy without exposing sensitive information, mitigating the risk of privacy breaches \parencite{drechsler_synthetic_2011, hu_advancing_2024}. Another benefit of synthetic data is its flexibility across various levels of detail and complexity, successfully tailoring to specific use cases and diverse features like those found in business surveys \parencite{kim_synthetic_2021}. 

The public and some government agencies remain hesitant to rely on or use synthetic data due to concerns over its accuracy, reliability, and potential biases. However, the US Census Bureau is making significant progress in addressing these concerns and is no stranger to synthetic microdata products \parencite{jarmin_expanding_2014, miranda_using_2016}. In May 2013, the partially synthetic SIPP Synthetic Beta product was publicly released, integrating person-level microdata from the Survey of Income \& Program Participation (SIPP) with administrative tax and benefit data \parencite{benedetto_creation_2013}. In 2018, the fully synthetic version was released \parencite{benedetto_creation_2018}. For business data in particular, the Synthetic Longitudinal Business Database (SynLBD) is the first establishment-level PUMS ever released by a US statistical agency, consisting of 21 million establishment records from all sectors over the span of 1976 – 2000 \parencite{kinney_towards_2011}. As pivotal as this release was, the SynLBD is still limited-use in the sense that it is listed as an “experimental synthetic data product” without geographic nor firm-level information and there is no guarantee analyses from the SynLBD reflect the analyses from the underlying restricted-use data \parencite{kinney_improving_2014}. Synthetic data overall has the potential to bridge the gap between the challenges of de-identification and re-identification of firm-level data when synthesized appropriately, giving rise to a privacy-preserving solution while maintaining data utility and integrity. 

There are numerous approaches for creating synthetic data without compromising the privacy of sensitive information. At a high level, synthetic data generation is essentially a task of fitting a generative model to a given dataset, generally accomplished by capturing the distribution of the original data in some fashion and then sampling another dataset from that same distribution. Naturally, these models introduce an approximation error when producing complex, high-dimensional, potentially sparse data, which will then be used to run other analyses or even train machine-learning (ML) models downstream for regression or classification tasks. Accuracy degrades at each step in the chain, so careful attention is devoted to developing a generative model and its evaluation and validation on the ground-truth restricted-use data. High-quality synthetic survey data should ideally be as similar to the original data as two uniform random partitions of the original data are to each other. Privacy-preserving data generators can be iteratively tested and safely improved using previously released public datasets of the target schema, where available. 

AI techniques play a crucial role in generating synthetic data by leveraging advanced algorithms to create artificial datasets that closely mimic the statistical properties and relationships present in real-world data. By learning and replicating the patterns and structures within the original data, AI-generated synthetic data ensures that key correlations and distributions are preserved. Modelling via classification and regression trees (CART) is a powerful yet intuitive ML technique for building decision trees that can effectively learn to predict values of variables based on the values of other variables in the data \parencite{breiman_classification_1984}. CART-based synthesizers consistently exhibit strong performance on tabular datasets \parencite{pathare2023} even when compared to complex deep learning models, such variational autoencoders (VAEs) and generative adversarial networks (GANs).  Decision trees are hierarchical structures that recursively split the data based on features to create nodes representing different decisions or values. Initial data partitions from the restricted-use microdata are channelled down into different paths of a tree, where hyperparameters are iteratively tuned in order to best branch off the population into low-entropy and self-similar groups with respect to a target variable. This leads to a tree structure that captures complex relationships and patterns found in the restricted-use data. Decision trees generate synthetic data by following the same splitting rules learned during the training process, producing synthetic samples that maintain the underlying patterns and distributions observed in the real data \parencite{eno_generating_2008, fletcher_decision_2020, freiman_data_2017}. 

CART-based synthesizers are used today at statistical agencies around the world. In 2016, the R \emph{synthpop} library was created as a component of the SYLLS project (SYnthetic Data Estimation for UK LongitudinaL Studies) from the UK Economic and Social Research Council to produce synthetic data tailored to the needs of individual research projects, namely the Scottish Longitudinal Study \parencite{nowok_utility_2015, nowok_providing_2017}. More recently, the R \emph{tidysynthesis} library from the Urban Institute and the US Internal Revenue Service was used to generate fully synthetic tax data in 2022 \parencite{williams_tidysynthesis_2022}. Since 2018, Knexus has been developing CenSyn for and in collaboration with the US Census Bureau and CenSyn has been configured for multiple groups and data products, like the full synthesis of the American Housing Survey, the American Community Survey (ACS), the ACS – Housing Unit Data, and the ACS – Group Quarters Data. At the Data Privacy Protection \& the Conduct of Applied Research Conference in May 2024, \cite{totty_statistical_2024} additionally showed that the error from CenSyn in a select set of estimated means and population sizes on the ACS data is smaller on average than measurement error or coverage error and similar in magnitude to non-response error \parencite{totty_statistical_2024}. 

Given decision trees' diverse capabilities in handling complexities commonly encountered in surveys conducted by national statistical organizations, we choose to implement CART modelling, specifically CenSyn and synthpop, for generating synthetic business microdata corresponding to the ABS. The resulting product can be used for educational purposes, testing code, testing analysis, or performing aggregate analysis on the population. Provided with proper tuning of tree parameters, the synthetic firms cannot be used for learning about specific respondents in the restricted dataset.

\section{Our Synthetic Data Framework}
CenSyn consists of both a Synthesizer (sequence of decision trees) and an Evaluator suite. The latter serves as a flexible set of tools for determining the quality of the privacy-preserving synthetic data from the Synthesizer, which in turn depends on many factors, such as data pre-preprocessing, encoding, dataset partitioning, and model parameters. As such, relative metrics for quantifying distributions between the original and synthetic data are required in order to compare different configurations within the Synthesizer. In particular, the k-marginal tool is a quickly computable distribution similarity metric based on randomly sampled k-marginals \parencite{sen2023}. Integrated within the CenSyn production-level software, this metric assists in tracking down deviations in the data distribution with detailed reporting and the ability to adjust the span and width of the marginals. This tool gives an easy, detailed score between 0 and 1000 that is invaluable for optimization and parameter fitting during the Synthesizer’s development. Specifically, a score of 0 implies no overlap whatsoever between the ground-truth and synthetic PUMS, while a score of 1000 implies perfectly matching density distributions. Empirically, a score of at least 970 is sought for a satisfactory synthetic PUMS.  Evaluation of synthetic data is a challenging, active, and vitally important topic of research across many fields. Metrics are necessary for the engineering and development of practical and usable synthetic data generators for large-scale business surveys like the ABS.

In addition to the Synthesizer and Evaluators, CenSyn efficiently performs privacy checks, consistency checks, handles weights, performs full or partial synthesis (synthesizing all or selected features), and preserves the sparsity of the original data. This last point is highly relevant for the ABS and its R\&D and innovation modules, where some firms participate in more than others, and hence increase the risk of compromising confidentiality. CenSyn’s configuration of pre-processing and sequence of decision trees capture hierarchical decision rules and can identify and represent rare events and sparse data within their branches to reduce overfitting and improve generalization. The final synthetic microdata product preserves the distribution of the original data, in all its diversity, while also ensuring difficulty in identifying any real record through its highly tuneable configuration. In an interactive manner between researchers, survey managers, and stakeholders, we first perform an initial round of evaluation and tuning to identify the poorest performing features in the synthetic data and modify the synthesis configuration to solve any obvious issues. Then, we communicate results from the Evaluators with stakeholders and survey managers and rely on their expertise to hypothesize why certain features are harder to model. From the insight gained, we explore solutions in a third phase to tune the Synthesizer and evaluate the effects. The second and third phases are iterated as necessary until a final product is achieved.

The synthpop package also comes equipped with both synthesis and evaluation tools. In order to test sensitivity to specific software configurations, we implement parallel design and development processes using synthpop. We leverage the evaluation metrics of both frameworks to compare performance and adjust settings for the CART models.

\section{SBO 2007 PUMS}
The long-term aim of this project is to produce a synthetic ABS PUMS via CART-based models. We are currently evaluating synthetic products and fine-tuning, but these results are unfortunately still restricted at the time of this submission. However, we configured CenSyn and synthpop to generate synthetic microdata based on the SBO 2007 PUMS, which is not restricted and is similar to the firm-level data collected in the ABS. Treating the SBO 2007 PUMS as the “ground-truth” or “original” data, we can more openly explore and present the relationships that are preserved in the resulting synthetic microdata products. In addition, insights discovered during this phase are carried into developing the ABS PUMS. 
The SBO collects data on gender, ethnicity, race, and veteran status of business owners, along with administrative data and business characteristics, like the year the business was established, exports, and employee information \parencite{us_census_bureau_2007_2007}. Specifically, the 2007 PUMS consists of over 2.1 million firms and 200 features with sector-level industry (2-digit NAICS code) and state-level geography (Federal Information Processing Standard (FIPS) state codes), enabling the creation of custom tables tailored to various user needs. 

\begin{figure}[ht]
    \centering
    \includegraphics[width=1\linewidth, trim={0 0.9in 0 0},clip]{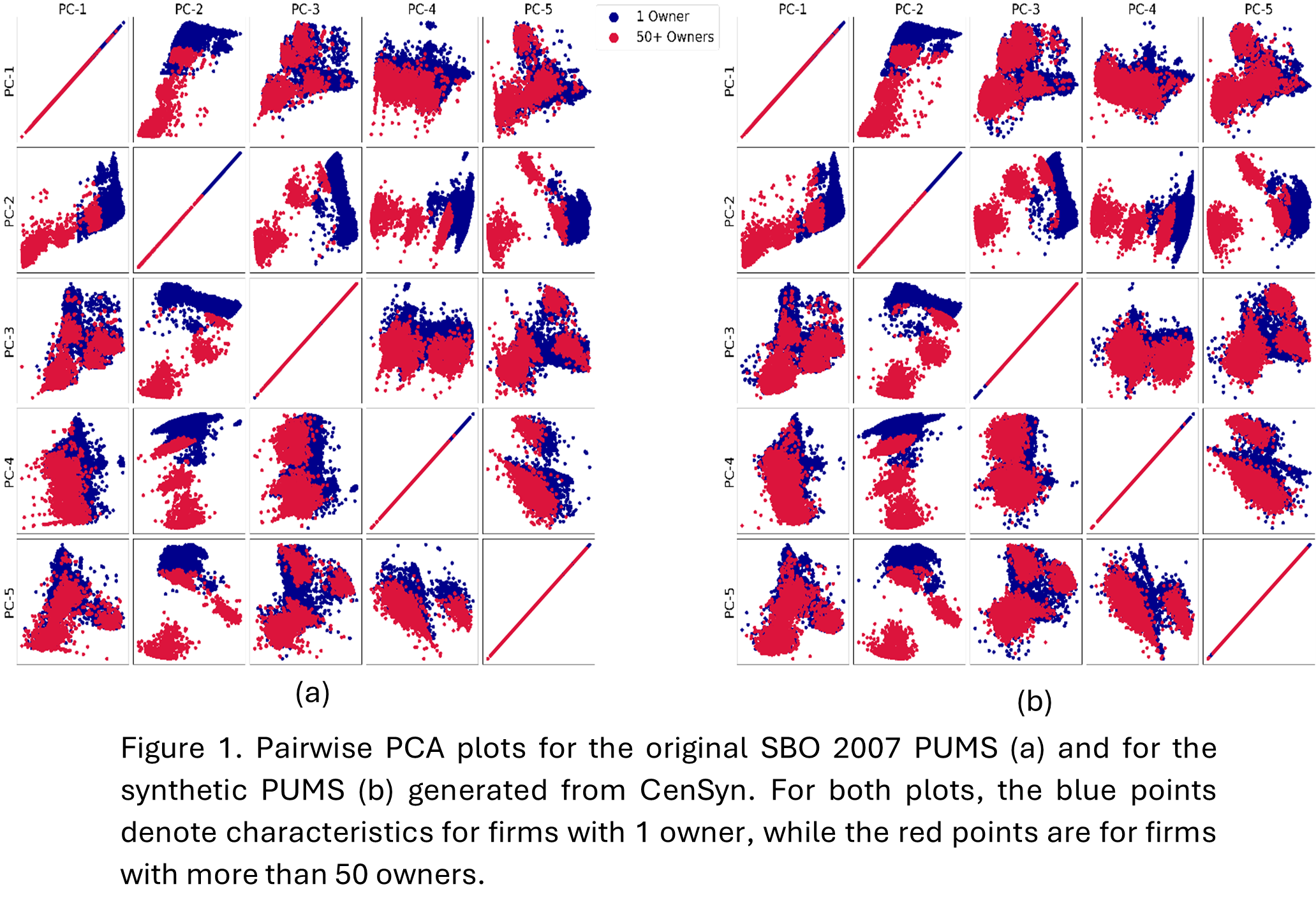}
    \caption{Pairwise PCA plots for the original SBO 2007 PUMS (a) and for the synthetic PUMS (b) generated from CenSyn. For both plots, the blue points denote characteristics for firms with 1 owner, while the red points are for firms with more than 50 owners.}
    \label{fig:1}
\end{figure}

Following the discussion process described in the previous section, the original 2007 PUMS is fed into the either synthesizer and the synthetic product is iteratively fine-tuned. The results presented here correspond to two synthetic PUMS of over 1 million synthetic firms each, roughly half the number of firms of the original PUMS (for computational feasibility) with 54 features. The Synthpop-synthesized dataset had a k-marginal score of 992 (with a 30\% baseline score of 994) and the Censyn-synthesized dataset also scored a 992 (with a 30\% baseline score of 994). Based on other metrics, the state feature (‘FIPST’) and revenue (‘RECEIPTS\_NOISY’) were the worst performing features. The former is likely due to the diversity in responses across the entirety of the original microdata, while the latter might be due to the noise additionally infused into the released 2007 PUMS to further protect the confidentiality of individuals and businesses (we do not anticipate similar complications in the internal ABS data). To visualize how well the synthetic PUMS captures the distributions of the original PUMS, we focus primarily on the pairwise PCA (principal component analysis) tool, adapted from the High-Level Group on the Modernization of Official Statistics Synthetic Data Challenge \parencite{task_sdnist_2023}. Briefly, PCA is a dimensionality reduction technique that transforms a large set of correlated variables into a smaller set of uncorrelated variables, called principal components (PCs), that capture the variance in the data \parencite{hotelling_analysis_1933, wold_principal_1987}. The goal is to represent the original data with the least number of PCs. In pairwise PCA, we first perform PCA and then plot pairwise combinations of the top 5 PCs into 2D scatter plots to visualize simplified distributions found in the data. If two datasets are similar, then their pairwise PCA plots will show agreement in the distribution shapes and dispersion patterns.

Therefore, computing the pairwise PCA for the original PUMS on the same features from the CenSyn synthetic PUMS and computing the pairwise PCA for the CenSyn PUMS yields Figures \ref{fig:1}(a) and \ref{fig:1}(b) separately. Each set of plots are color-coded to denote firms with 1 owner and firms with 50 or more owners – business characteristics of interest that are common and rare, respectively, in the original PUMS. We see high correlations between the original and synthetic data among the top 3 PCs, matching the distribution shapes and patterns. Table \ref{tab:pc_percentages} presents the percentages of variance explained by each of the selected components and confirms that the top 5 PCs for the synthetic PUMS contribute just as much as the top 5 PCs for the original PUMS. The PCs from both PUMS account for a quarter of the total variance in the data, with PC-1 contributing about 14\% and PC-5 nearly 1\%. This implies that nearly all 54 chosen features for the synthetic PUMS are vital enough to practically capture the equivalent level of information as the same 54 features in the original PUMS with over twice the number of firms. We find similar results when comparing the synthpop synthetic PUMS. 

\begin{table}[ht]
\centering
\begin{tabular}{lccccc}
\hline
 & PC-1 & PC-2 & PC-3 & PC-4 & PC-5 \\
\hline
SBO 2007 PUMS & 13.78\% & 7.12\% & 3.12\% & 1.55\% & 1.35\% \\
CenSyn PUMS   & 13.78\% & 7.09\% & 3.11\% & 1.53\% & 1.36\% \\
\hline
\end{tabular}
\caption{Percentages of variance explained by each of the top PCs corresponding to pairwise PCA in Figure \ref{fig:1}}
\label{tab:pc_percentages}
\end{table}

\begin{figure}[ht]
    \centering
    \includegraphics[width=1\linewidth, trim={0 1.7in 0 0},clip]{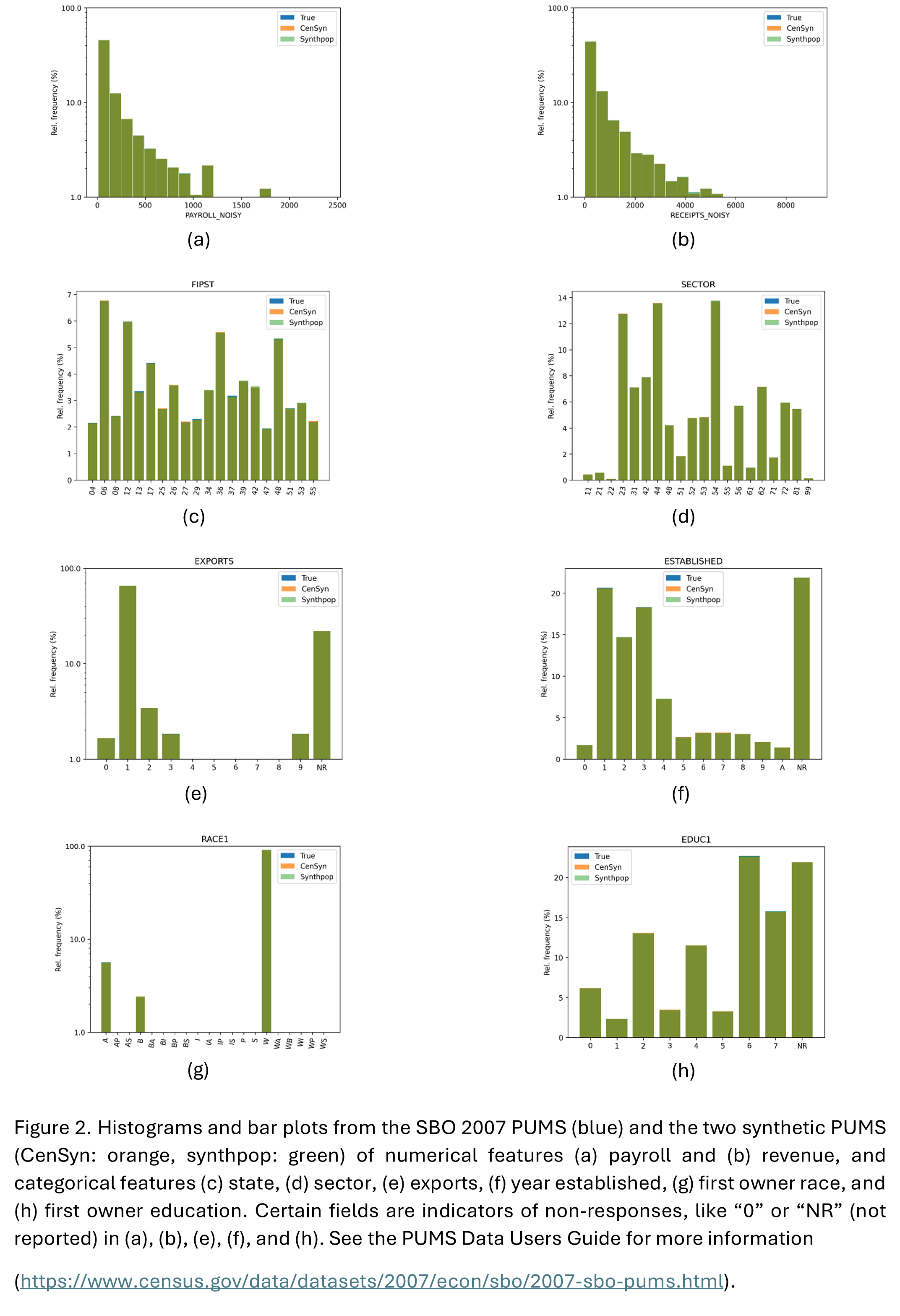}
    \caption{Histograms and bar plots from the SBO 2007 PUMS (blue) and the two synthetic PUMS (CenSyn: orange, synthpop: green) of numerical features (a) payroll and (b) revenue, and categorical features (c) state, (d) sector, (e) exports, (f) year established, (g) first owner race, and (h) first owner education. Certain fields are indicators of non-responses, like “0” or “NR” (not reported) in (a), (b), (e), (f), and (h). See the PUMS Data Users Guide for more information (\href{https://www.census.gov/data/datasets/2007/econ/sbo/2007-sbo-pums.html}{https://www.census.gov/data/datasets/2007/econ/sbo/2007-sbo-pums.html}).
}
    \label{fig:2}
\end{figure}

\begin{figure}
    \centering
    \includegraphics[width=1\linewidth, trim={0 2.1in 0 0},clip]{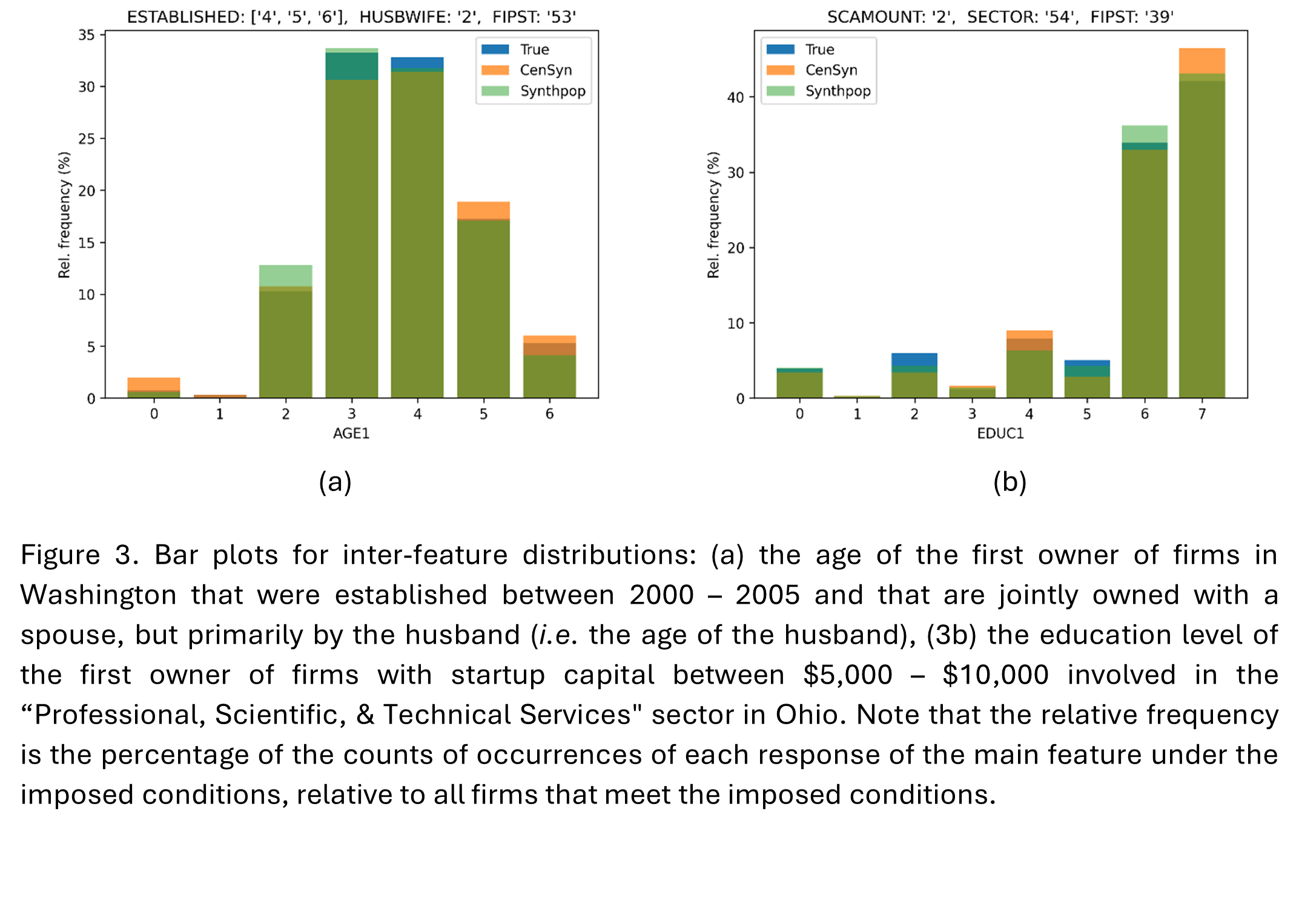}
    \caption{Bar plots for inter-feature distributions: (a) the age of the first owner of firms in Washington that were established between 2000 – 2005 and that are jointly owned with a spouse, but primarily by the husband (i.e. the age of the husband), (b) the education level of the first owner of firms with startup capital between \$5,000 – \$10,000 involved in the “Professional, Scientific, \& Technical Services" sector in Ohio. Note that the relative frequency is the percentage of the counts of occurrences of each response of the main feature under the imposed conditions, relative to all firms that meet the imposed conditions.}
    \label{fig:3}
\end{figure}

In addition to pairwise PCA, we present standard results by directly comparing how well individual feature distributions are preserved in both synthetic PUMS. Figure \ref{fig:2} shows histograms and bar plots for payroll, revenue, state, sector, exports, year established, first owner race, and first owner education. In contrast, Figure \ref{fig:3} depicts how well inter-feature distributions are preserved in a given US state. Precisely, in Figure \ref{fig:3}(a), we plot the age of the first owner of firms in Washington that were established between 2000 – 2005 and that are jointly owned with a spouse, but primarily by the husband. Figure \ref{fig:3}(b) shows the education level of the first owner of firms with startup capital between \$5,000 – \$10,000 involved in the “Professional, Scientific, \& Technical Services" sector in Ohio. These quantitative and graphical evaluations showcase that the synthetic PUMS accurately learned the characteristics of the original SBO 2007 PUMS, with some deviations when studying multi-variable dependencies. 

\FloatBarrier
\section{Replicating an Analysis in \emph{Small Business Economics} Using Synthetic Data}
Reproducing the first moments of the true data as demonstrated in Figures \ref{fig:2} and \ref{fig:3} provides some assurance of the eventual ability of de-identified synthetic data to substitute for restricted data. However, to be useful for scientific exploration, synthetic data must adequately preserve the cross-product moments of restricted data that are the focus of social science research. Replicating “Transnational activities of immigrant-owned firms and their performances in the USA,” a highly impactful article by \textcite{wang2015} using synthetic versions of the 2007 SBO PUMS, provides several hard tests. First, the analysis focuses on relatively rare phenomena such as firms with operations overseas, firms that export, and firms that outsource. Second, the estimation includes industry fixed effects that expand the number of coefficient estimates that may differ in sign and significance from the true data. And third, the analysis includes both binary and continuous dependent variables estimated with logistic regression and ordinary least squares, respectively. All told, 26 independent variables are in the limited dependent variable and continuous variable models that provide real-world tests of the synthetic data.

The goal is not to reproduce the \textcite{wang2015} results exactly but rather to compare results from a similar specification from the 2007 SBO PUMS and the two synthesized datasets that used this PUMS as training data. The main difference with the original analysis is that the non-employer businesses are not included in the sample as the exercise is intended to inform synthetic data production for the Annual Business Survey that includes only employer businesses. The other difference is that three of the control variables used in \textcite{wang2015} had not been included in the set of features to be simulated: whether the business is home-based, and whether the business sold to businesses or consumers. The home-based business question is more relevant to non-employer businesses. Type of customer questions are included in ABS and it will be important to include these in the eventual synthetic public use file. However, the main interest in this exercise is in testing the verisimilitude of the synthetic datasets relative to the observational data.

The central question investigated is whether firms owned by immigrants are more likely to be engaged in the transnational activities of operating a branch overseas, exporting, and foreign outsourcing. The economic impacts of transnational activities are also investigated with respect to sales per employee and payroll per employee. Table \ref{tab:pums_comparison} provides descriptive statistics for the dependent variables and the independent variables of main interest across the three datasets. As demonstrated in Figures \ref{fig:2} and \ref{fig:3}, the means from the synthetic datasets are very similar to the observed 2007 SBO dataset. Given the rarity of some of the features, similarity to the third decimal place for many of the estimates is encouraging regarding the eventual application to innovation data that often capture rare phenomena. Most of the variable names are self-explanatory with the exception of Transnational that indicates any international activity in the form of exporting, outsourcing, or having operations overseas.

\begin{table}[ht]
\centering
\begin{tabular}{lccc}
\hline
\textbf{Variable Name} & \textbf{2007 SBO PUMS} & \textbf{CenSyn PUMS} & \textbf{synthpop PUMS} \\
\hline
International Branch & 0.0133 & 0.0126 & 0.0131 \\
Exporting & 0.1150 & 0.1161 & 0.1156 \\
International Outsourcing & 0.0164 & 0.0164 & 0.0163 \\
Ln Sales per Employee & 4.7544 & 4.7509 & 4.7536 \\
Ln Payroll per Employee & 3.3256 & 3.3246 & 3.3259 \\
Immigrant & 0.1143 & 0.1150 & 0.1135 \\
Immigrant x Family Business & 0.0388 & 0.0388 & 0.0379 \\
Transnational & 0.1272 & 0.1286 & 0.1280 \\
Transnational x Immigrant & 0.0188 & 0.0177 & 0.0184 \\
\hline
\end{tabular}
\caption{Descriptive statistics of dependent and selected independent variables}
\label{tab:pums_comparison}
\end{table}

Testing whether the synthetic datasets are able to preserve the cross-product moments of the observational data can be done by replicating the econometric models used in \textcite{wang2015}. Table \ref{tab:3} provides a summary of the main independent variables of interest examining the association of immigrant status with transnational activity. The table contains the 95th percent confidence interval of the estimated odds ratios. Confidence intervals of odds ratios have three distinct advantages: 1) an odds ratio is easily interpreted as the change in odds given a one-unit change in the independent variable; 2) statistical significance is easily determined by whether the entire interval is above 1 (indicating a statistically significant positive log odds estimate) or below 1 (indicating a statistically significant negative log odds estimate); and 3) statistically significant differences comparing estimates across datasets are evident when confidence intervals do not intersect (A detailed description of the method of overlapping confidence intervals and its comparison with the standard methods for testing of a hypothesis is given in \textcite{schenker2001}). Qualitatively, the results from all three datasets are identical. Immigrant ownership is positively associated with all three transnational activities in all three datasets. However, immigrant-owned family businesses are negatively associated with exporting in all three datasets but not statistically significant in any of the International Branch or International Outsourcing equations. Quantitatively, none of the immigrant-owned family business estimates are statistically different but some of the immigrant estimates are. The immigrant CenSyn estimates are statistically lower than the 2007 SBO estimates in the International Branch and Exporting equations but not statistically different in the International Outsourcing equation. The converse holds for the synthpop estimates that are not statistically different from the 2007 SBO estimates in the International Branch or Exporting equations but statistically lower in the International Outsourcing equation.

\begin{table}[ht]
\centering
\scriptsize
\begin{tabular}{l|cccc|cccc|cccc}
\hline
 & \multicolumn{4}{c|}{\textbf{International Branch}} & \multicolumn{4}{c|}{\textbf{Exporting}} & \multicolumn{4}{c}{\textbf{International Outsourcing}} \\
 \textbf{Dataset} & \multicolumn{2}{c}{Immigrant} & \multicolumn{2}{c|}{Immigrant*Family} & \multicolumn{2}{c}{Immigrant} & \multicolumn{2}{c|}{Immigrant*Family} & \multicolumn{2}{c}{Immigrant} & \multicolumn{2}{c}{Immigrant*Family} \\
\hline
2007 SBO & 2.078 & 2.385 & 0.858 & 1.055 & 1.739 & 1.851 & 0.835 & 0.914 & 1.933 & 2.191 & 0.974 & 1.161 \\
CenSyn SBO & 1.604 & 1.860 & 0.905 & 1.090 & 1.422 & 1.513 & 0.848 & 0.929 & 1.756 & 1.995 & 0.862 & 1.041 \\
synthpop SBO & 1.976 & 2.279 & 0.910 & 1.118 & 1.666 & 1.772 & 0.804 & 0.881 & 1.620 & 1.847 & 0.876 & 1.056 \\
\hline
\end{tabular}
\caption{95\% Confidence Interval of Odds Ratios for Main Variables of Interest from \textcite{wang2015}}
\label{tab:3}
\end{table}

Moving to the OLS regression summary results in Table \ref{tab:4}, we see that the qualitative results are again identical with one exception: the Immigrant estimate in the Sales per Employee equation is not statistically significant when using the CenSyn dataset but is using the observational dataset or the synthpop dataset. Quantitatively, the large effect of Transnational business on Sales per Employee and Payroll per Employee is larger using the observational dataset than when using either synthetic dataset. However, this is not the case when looking at immigrant owned transnational businesses. None of the estimates are statistically significant in the Sales per Employee equations. In the Payroll per Employee equation none of the estimates using the various datasets are statistically different.  One might expect observational data to regularly estimate larger effects as outliers and extreme values are more likely than in synthetic data. This appears to be the case for the more prevalent Transnational feature (characterizing a 0.1272 share of firms from Table \ref{tab:pums_comparison}) but not evident for the much rarer immigrant owned transnational business (0.0188). 

\begin{table}[ht]
\centering
\scriptsize
\begin{tabular}{l|cccccc|cccccc}
\hline
 & \multicolumn{6}{c|}{\textbf{Ln Sales per Employee}} & \multicolumn{6}{c}{\textbf{Ln Payroll per Employee}} \\
\textbf{Dataset} & \multicolumn{2}{c}{Transnational} & \multicolumn{2}{c}{Immigrant} & \multicolumn{2}{c|}{Transnat Immigrant} & \multicolumn{2}{c}{Transnational} & \multicolumn{2}{c}{Immigrant} & \multicolumn{2}{c}{Transnat Immigrant} \\
\hline
2007 SBO & 0.3375 & 0.3538 & 0.0280 & 0.0472 & -0.0206 & 0.0198 & 0.2554 & 0.2684 & -0.0774 & -0.0619 & 0.0162 & 0.0486 \\
CenSyn SBO & 0.2959 & 0.3120 & -0.0111 & 0.0076 & -0.0357 & 0.0058 & 0.2125 & 0.2255 & -0.0966 & -0.0814 & 0.0204 & 0.0539 \\
synthpop SBO & 0.2983 & 0.3144 & 0.0094 & 0.0285 & -0.0257 & 0.0151 & 0.2185 & 0.2315 & -0.0691 & -0.0535 & 0.0187 & 0.0517 \\
\hline
\end{tabular}
\caption{95\% Confidence Intervals of Coefficient Estimates for Main Variables of Interest from \textcite{wang2015}}
\label{tab:4}
\end{table}

Further tests of the verisimilitude of the synthetic datasets are provided in the Appendix Tables \ref{tab:logit_models_full} and \ref{tab:sales_payroll_models} that provide the full set of regression results with all industry, firm, and demographic controls. The most revealing set of coefficient estimates are those used to control for industry in Table \ref{tab:logit_models_full}. Summary industries differ significantly regarding their ability and proclivity for transnational activity. The observational dataset picks this up well with difficult-to-export industries such as Construction, Social Services, and Personal Services characterized by large negative estimates and highly tradable industries such as Wholesale, Transportation, and Information having positive and statistically significant estimates (Manufacturing is the excluded category). These phenomena are picked up in the synthetic datasets with generally similar magnitudes. Given the central role that industry fixed effects play in nearly all econometric estimates using business establishment data, verisimilitude with respect to this dimension is particularly reassuring.

\section{Discussion}
The ability of autoencoding software to produce synthetic facsimiles of observational data that preserve cross-product moments from the original data is the critical determinant of its usefulness as a replacement of or supplement to restricted confidential data. The replication of a highly impactful article from Small Business Economics using synthetic data produced from the 2007 SBO public-use file  confirms a high degree of verisimilitude of the synthetic data. Most importantly, the synthetic datasets were able to capture relatively rare phenomena with a high degree of precision along with the cross-product moments of these rare phenomena with outcomes of interest. The ability to preserve cross-product moments extends to a long list of control variables, including industry fixed effects, reinforcing confidence in synthetic data as a productive tool to explore phenomena of interest while controlling for potential confounds. However, synthetic data may be at a disadvantage in those cases where the exact magnitude of a coefficient estimate is critical from a policy or scientific perspective. Magnitudes were statistically different between the observed and synthetic data in the replication exercise when examining the association between transnational activities and sales or payroll per employee. This finding is not surprising as extreme values in the observed data are less likely to be reproduced in the synthetic data.

Given the inherently stochastic nature of observational data, synthetic data may provide a preferable environment to explore phenomena of interest. Extreme values in a sample may be influential in selecting a particular specification after repeated specification testing. If the same data are used for both specification and hypothesis testing, the seemingly strong empirical results may not be robust when estimated with a new sample. Synthetic data presents the possibility of repeated specification testing on data that reproduce the cross-product moments of the observed data, providing the information needed to produce a detailed pre-analysis plan. The \emph{de novo} estimation of the confidential observed data using this detailed pre-analysis plan preserves the validity of statistical tests from econometric software that assumes a single test is being performed \parencite{goldberger1991}. The use of synthetic data for specification testing and the corresponding confidential data for hypothesis testing has the potential to reduce the probability of false discovery that is a principal contributing factor to the replicability crisis in applied economics research \parencite{ferraro2023}. The probability of false discovery in the economics of innovation research is arguably even higher given innovation phenomena concentrating in the right tail of statistical distributions that may only be detected using quantile regression techniques \parencite{coad2008}. Determining the best quantile for detecting innovation phenomena using a synthetic dataset for specification testing that is then confirmed in a single hypothesis test using confidential observational data can drastically reduce researcher degrees of freedom, thus reducing the probability of false discovery \parencite{breznau2022observing}.  

\section{Conclusion, Limitations, and Future Research}
\begin{figure}
    \centering
    \includegraphics[width=1\linewidth, trim={0 1.2in 0 0},clip]{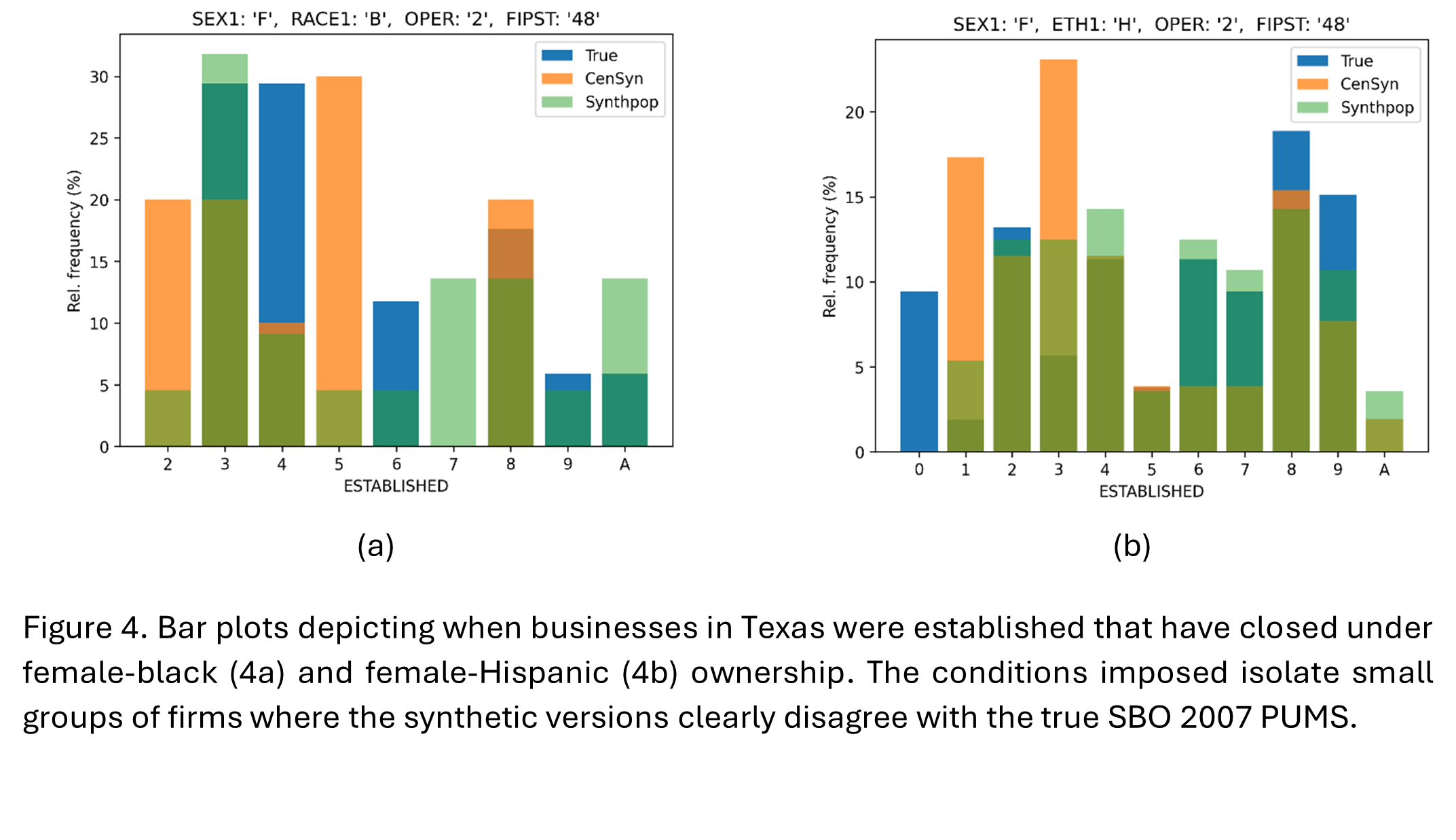}
    \caption{Bar plots depicting when businesses in Texas were established that have closed under female-black (a) and female-Hispanic (b) ownership. The conditions imposed isolate small groups of firms where the synthetic versions clearly disagree with the true SBO 2007 PUMS.}
    \label{fig:4}
\end{figure}

The replication exercise of the \textcite{wang2015} analysis using 2007 SBO employer data and synthetic data sets is generally encouraging. However, an attempt to replicate an analysis published in the \emph{American Economic Review} on the failure rate of firms in their first year of operation \parencite{mora2014} that also used the 2007 SBO public use file did not produce identical qualitative results using synthetic data. The possible reasons for this are instructive of potential limitations of synthetic data. First, restricting analysis to firms in their first year of operation in 2007 reduced the sample size by more than a factor of ten. The large sample size of government datasets such as SBO and ABS help to mitigate the introduction of errors that guard against re-identification. Limiting analysis to a small subset of a synthetic dataset may reduce verisimilitude available in analysis using the full sample. This possibility will be a topic of future research given the strong research interest in R\&D-performing microbusinesses in ABS that also make up a small share of ABS. Second, analysis based on the interaction of two or more synthesized variables may be susceptible to the amplification of errors (see Figure \ref{fig:4}). Two central variables in \textcite{mora2014} are the interaction of the female owned feature with both the black-owned and Hispanic-owned features. While these interactions represent between 1.3\% to 1.5\% of firms in the observed data, the percentages in both synthetic datasets were closer to 1\%. Both of these points are evident in Figure 4. Finally, some phenomena being studied may be inherently more difficult to model, making replication with synthetic data more challenging. The replication presented here of transnational activity can be thought of as a right tail event similar to many innovative phenomena. In contrast, the phenomenon of interest in Mora and Dávila (2014) of firms that go out of business in their first year of operation could be classified as a left tail event. The factors explaining success may be more systemic than the factors associated with failure, similar to famous opening line by \textcite{tolstoy1995}: ``All happy families are alike; every unhappy family is unhappy in its own way.''

Another issue not yet addressed is the ability to integrate geographic factors into the analysis of synthetic data. Geographical analysis is currently done with restricted access ABS by linking the county-level geocodes available in the dataset to public or confidential data on county characteristics related to such things as digital infrastructure \parencite{han2023}, human capital endowments, or social capital \parencite{han2025}. The combination of industry membership and county location drastically reduces the number of candidate synthetic firms that might resemble respondents in the observed dataset. One strategy being investigated to mitigate re-identification risk is clustering NAICS codes and geography variables that address distribution sparsity yet guarantee that the synthetic PUMS reflects the industry landscape as best as possible. Determining the optimal level of geographic detail for data release is essential, so we will explore factors such as urban-rural distinctions via RUCC (Rural-Urban Continuum Codes) and MSAs to provide meaningful insights into regional trends and patterns. We will also explore the possibility of releasing several PUMS of different levels of details for different needs. The re-identification risks must be further carefully studied in this case, where perhaps one PUMS carries 2-digit NAICS but county-level information, while a second PUMS dives into 6-digit NAICS but only at a clustered state level. 

PUMS from business surveys play a critical role in facilitating public access to government-funded data for research and decision-making across academic, governmental, and industrial sectors; however, the absence of these PUMS has posed challenges in meeting analytical demands in a timely manner while safeguarding privacy. We emphasize the need for innovative solutions that balance data utility with privacy protection in the realm of business microdata. We introduce ongoing conversations in developing a novel approach utilizing synthetic data generation powered by AI to address the shortcomings of traditional de-identification methods for business survey microdata, specifically those collected in the ABS. By leveraging decision trees like those configured in CenSyn and synthpop, we propose a privacy-preserving solution that generates synthetic microdata representative of the restricted-use microdata's statistical properties while mitigating re-identification risks. While we continue to test on the SBO 2007 PUMS and internal ABS microdata, we aim to communicate with current and potential ABS users in an iterative manner to gain additional insights before releasing a final ABS PUMS cleared by the Census Disclosure Review Board. 

\FloatBarrier
\printbibliography

\section*{Declarations}
Competing Interests: Cisneros Paz, Wojan, Williams, Ozawa, Chew, Janda, and Navarro have no competing interests to declare that are relevant to the content of this article. Floyd, Task, and Streat are employees of Knexus Research LLC. Knexus developed the CenSyn synthetic data generator used in this paper for the United States Census Bureau.

Funding:  Funding for this project came from National Center for Science and Engineering Statistics ADC Base Agreement No. 2023-605.  Contributions from Jorge Cisneros Pas and Timothy Wojan were performed under an appointment administered by the Oak Ridge Institute for Science and Education (ORISE) and supported in part by the National Science Foundation (NSF), National Center for Science and Engineering Statistics (NCSES) by the Oak Ridge Institute for Science and Education (ORISE) for the Department of Energy (DOE). ORISE is managed by Oak Ridge Associated Universities (ORAU) under DOE contract number DE-SC0014664. This paper provides results of exploratory research sponsored by the National Center for Science \& Engineering Statistics (NCSES) within the United States (US) National Science Foundation (NSF). This information is being shared to inform interested parties of ongoing activities and to encourage further discussion. All opinions expressed are the authors’ and do not necessarily reflect the policies and views of NCSES, NSF, DOE, ORAU, or ORISE.

Data and Code Availability: All data used in this analysis are available in \href{https://gitlab.com/KnexusPublic/ncses-abs-syntheticdata-paper}{https://gitlab.com/KnexusPublic/ncses-abs-syntheticdata-paper}.  The code used in the replication exercise is provided in the Supplementary Information file along with releasable code related to the creation of synthetic data using R \emph{synthpop}. CenSyn is proprietary software written by Knexus Research LLC and provided to the United States Census Bureau. CenSyn was made available for use on this project but its licensing and restrictions do not allow for wider public distribution.

\appendix
\setcounter{table}{0}
\renewcommand{\thetable}{\Alph{section}\arabic{table}}
\section{Appendix Tables}

\begin{table}[ht]
\centering
\scriptsize
\begin{tabular}{l|rrr|rrr|rrr}
\hline
\textbf{Parameter} & \multicolumn{3}{c|}{\textbf{International}} & \multicolumn{3}{c|}{\textbf{Exporting}} & \multicolumn{3}{c}{\textbf{Outsourcing}} \\
(p-value) & 2007 SBO & CenSyn SBO & synthpop SBO & 2007 SBO & CenSyn SBO & synthpop SBO & 2007 SBO & CenSyn SBO & synthpop SBO \\
\hline
Intercept        & -4.3079 & -4.3838 & -4.4552 & -2.4128 & -2.3321 & -2.3769 & -4.5353 & -4.3707 & -4.4408 \\
          & <.0001 & <.0001 & <.0001 & <.0001 & <.0001 & <.0001 & <.0001 & <.0001 & <.0001 \\
\hline immigrant        & 0.8002 & 0.5467 & 0.7524 & 0.5844 & 0.3832 & 0.5414 & 0.7216 & 0.6267 & 0.5480 \\
          & <.0001 & <.0001 & <.0001 & <.0001 & <.0001 & <.0001 & <.0001 & <.0001 & <.0001 \\
\hline immig\_family    & -0.0499 & 0.0107 & 0.00815 & -0.1356 & -0.1193 & -0.1724 & 0.0611 & -0.0538 & -0.0394 \\
          & 0.3421 & 0.8489 & 0.8767 & <.0001 & <.0001 & <.0001 & 0.1721 & 0.2635 & 0.4087 \\
\hline black            & -0.2588 & -0.2685 & -0.0809 & -0.7183 & -0.3114 & -0.2845 & -0.5762 & -0.1594 & -0.1881 \\
          & 0.0039 & 0.0023 & 0.3179 & <.0001 & <.0001 & <.0001 & <.0001 & 0.0255 & 0.0123 \\
\hline hispanic         & 0.2154 & 0.0760 & 0.0574 & 0.0222 & -0.0546 & 0.0937 & 0.1775 & -0.0868 & 0.2189 \\
          & <.0001 & 0.1392 & 0.2491 & 0.2862 & 0.0079 & <.0001 & <.0001 & 0.0687 & <.0001 \\
\hline asian            & 0.2352 & 0.1053 & 0.0574 & -0.2489 & -0.1568 & -0.2231 & 0.4779 & 0.2556 & 0.4567 \\
          & <.0001 & 0.0230 & 0.2000 & <.0001 & <.0001 & <.0001 & <.0001 & <.0001 & <.0001 \\
\hline mixedrace        & 0.4002 & 0.2495 & 0.2366 & -0.1235 & -0.0950 & -0.0481 & 0.3023 & 0.2117 & 0.1762 \\
          & 0.0005 & 0.0340 & 0.0409 & 0.0204 & 0.0568 & 0.3334 & 0.0045 & 0.0441 & 0.1027 \\
\hline family           & 0.0305 & 0.1314 & 0.1946 & 0.3124 & 0.3457 & 0.2924 & 0.1345 & 0.2126 & 0.2059 \\
          & 0.2594 & <.0001 & <.0001 & <.0001 & <.0001 & <.0001 & <.0001 & <.0001 & <.0001 \\
\hline female           & -0.5257 & -0.2755 & -0.1873 & -0.1749 & -0.1588 & -0.1049 & -0.3129 & -0.2030 & -0.1214 \\
          & <.0001 & <.0001 & <.0001 & <.0001 & <.0001 & <.0001 & <.0001 & <.0001 & <.0001 \\
\hline genderequal      & -0.3707 & -0.3789 & -0.3512 & -0.3317 & -0.3373 & -0.3171 & -0.2805 & -0.3236 & -0.2812 \\
          & <.0001 & <.0001 & <.0001 & <.0001 & <.0001 & <.0001 & <.0001 & <.0001 & <.0001 \\
\hline age35\_54        & -0.4101 & -0.1627 & -0.1851 & 0.1178 & 0.1239 & 0.1136 & -0.1284 & 0.0807 & 0.0219 \\
          & <.0001 & <.0001 & <.0001 & <.0001 & <.0001 & <.0001 & <.0001 & 0.0157 & 0.5177 \\
\hline agegt55          & -0.3762 & -0.1443 & -0.1082 & 0.1699 & 0.1440 & 0.1434 & -0.2348 & -0.0215 & 0.0208 \\
          & <.0001 & <.0001 & 0.0037 & <.0001 & <.0001 & <.0001 & <.0001 & 0.5309 & 0.5489 \\
\hline college          & 0.5583 & 0.3490 & 0.3624 & 0.3355 & 0.2187 & 0.2837 & 0.5512 & 0.2626 & 0.2881 \\
          & <.0001 & <.0001 & <.0001 & <.0001 & <.0001 & <.0001 & <.0001 & <.0001 & <.0001 \\
\hline est80\_89        & -0.1714 & -0.1174 & -0.1210 & -0.0976 & -0.1266 & -0.1054 & -0.00166 & -0.0985 & -0.0469 \\
          & <.0001 & <.0001 & <.0001 & <.0001 & <.0001 & <.0001 & 0.9514 & 0.0001 & 0.0720 \\
\hline est90\_99        & -0.1711 & -0.2251 & -0.1836 & -0.2683 & -0.2435 & -0.2878 & 0.0669 & -0.1375 & -0.0369 \\
          & <.0001 & <.0001 & <.0001 & <.0001 & <.0001 & <.0001 & 0.0093 & <.0001 & 0.1417 \\
\hline est00\_07        & -0.4198 & -0.4794 & -0.4547 & -0.5705 & -0.4445 & -0.5266 & -0.0046 & -0.3056 & -0.2450 \\
          & <.0001 & <.0001 & <.0001 & <.0001 & <.0001 & <.0001 & 0.8613 & <.0001 & <.0001 \\
\hline ecomm            & 1.2000 & 1.0341 & 1.1025 & 1.4853 & 1.4274 & 1.4804 & 1.1851 & 1.0569 & 1.0738 \\
          & <.0001 & <.0001 & <.0001 & <.0001 & <.0001 & <.0001 & <.0001 & <.0001 & <.0001 \\
\hline agric            & -0.1120 & -0.2453 & -0.3211 & -0.0430 & -0.0705 & -0.0636 & -1.4369 & -1.2532 & -1.5334 \\
          & 0.5442 & 0.2068 & 0.1048 & 0.4788 & 0.2415 & 0.2957 & <.0001 & <.0001 & <.0001 \\
\hline constr           & -1.2455 & -1.1969 & -1.2418 & -1.6809 & -1.6872 & -1.6652 & -1.6039 & -1.7058 & -1.6619 \\
          & <.0001 & <.0001 & <.0001 & <.0001 & <.0001 & <.0001 & <.0001 & <.0001 & <.0001 \\
\hline whlsle           & 0.5912 & 0.5906 & 0.5467 & 1.1949 & 1.2106 & 1.2154 & 0.4205 & 0.3825 & 0.3879 \\
          & <.0001 & <.0001 & <.0001 & <.0001 & <.0001 & <.0001 & <.0001 & <.0001 & <.0001 \\
\hline retail           & -0.4377 & -0.4803 & -0.4657 & -0.1270 & -0.1226 & -0.1046 & -0.8043 & -0.8035 & -0.7722 \\
          & <.0001 & <.0001 & <.0001 & <.0001 & <.0001 & <.0001 & <.0001 & <.0001 & <.0001 \\
\hline trans            & 0.4285 & 0.4125 & 0.3858 & 0.3175 & 0.2767 & 0.2966 & 0.3396 & 0.2480 & 0.2235 \\
          & <.0001 & <.0001 & <.0001 & <.0001 & <.0001 & <.0001 & <.0001 & <.0001 & <.0001 \\
\hline info             & 0.6304 & 0.7317 & 0.5574 & 0.4448 & 0.4712 & 0.4102 & 0.9771 & 1.1581 & 1.0296 \\
          & <.0001 & <.0001 & <.0001 & <.0001 & <.0001 & <.0001 & <.0001 & <.0001 & <.0001 \\
\hline profsvcs         & 0.2630 & 0.3315 & 0.3023 & -0.1921 & -0.1687 & -0.1912 & 0.4878 & 0.6269 & 0.5388 \\
          & <.0001 & <.0001 & <.0001 & <.0001 & <.0001 & <.0001 & <.0001 & <.0001 & <.0001 \\
\hline socsvcs          & -1.7206 & -1.6616 & -1.7521 & -2.3777 & -2.2616 & -2.4076 & -1.3578 & -1.1374 & -1.3373 \\
          & <.0001 & <.0001 & <.0001 & <.0001 & <.0001 & <.0001 & <.0001 & <.0001 & <.0001 \\
\hline persvcs          & -0.8937 & -0.9599 & -0.9692 & -0.6922 & -0.7160 & -0.7146 & -1.4463 & -1.5963 & -1.5335 \\
          & <.0001 & <.0001 & <.0001 & <.0001 & <.0001 & <.0001 & <.0001 & <.0001 & <.0001 \\
\hline
\end{tabular}
\caption{Logistic Regression with Full Set of Controls Corresponding to Table \ref{tab:3}}
\label{tab:logit_models_full}
\end{table}

\begin{table}[ht]
\centering
\scriptsize
\begin{tabular}{l|rrr|rrr}
\hline
\textbf{Variable} & \multicolumn{3}{c|}{\textbf{Sales per Employee}} & \multicolumn{3}{c}{\textbf{Payroll per Employee}} \\
(p-value) & 2007 SBO & CenSyn SBO & synthpop SBO & 2007 SBO & CenSyn SBO & synthpop SBO \\
\hline
Intercept      & 4.39882 & 4.40034 & 4.39472 & 3.12625 & 3.13687 & 3.14197 \\
        & <.0001 & <.0001 & <.0001 & <.0001 & <.0001 & <.0001 \\
\hline transnat       & 0.34563 & 0.30394 & 0.30635 & 0.26193 & 0.21901 & 0.22497 \\
        & <.0001 & <.0001 & <.0001 & <.0001 & <.0001 & <.0001 \\
\hline immigrant      & 0.03761 & -0.00172 & 0.01898 & -0.06965 & -0.08899 & -0.06130 \\
        & <.0001 & 0.7190 & <.0001 & <.0001 & <.0001 & <.0001 \\
\hline trans\_immig   & -0.00040 & -0.01494 & -0.00529 & 0.03242 & 0.03714 & 0.03518 \\
        & 0.9688 & 0.1586 & 0.6117 & <.0001 & <.0001 & <.0001 \\
\hline black          & -0.29941 & -0.22136 & -0.18284 & -0.25062 & -0.16036 & -0.16082 \\
        & <.0001 & <.0001 & <.0001 & <.0001 & <.0001 & <.0001 \\
\hline hispanic       & -0.09790 & -0.09733 & -0.07052 & -0.08301 & -0.09066 & -0.06699 \\
        & <.0001 & <.0001 & <.0001 & <.0001 & <.0001 & <.0001 \\
\hline asian          & -0.04208 & -0.07563 & -0.04097 & -0.14748 & -0.11579 & -0.12207 \\
        & <.0001 & <.0001 & <.0001 & <.0001 & <.0001 & <.0001 \\
\hline mixedrace      & -0.02642 & -0.05787 & -0.03761 & -0.07716 & -0.05743 & -0.07422 \\
        & 0.0757 & <.0001 & 0.0110 & <.0001 & <.0001 & <.0001 \\
\hline family         & 0.08839 & 0.10418 & 0.09527 & 0.07566 & 0.09237 & 0.08945 \\
        & <.0001 & <.0001 & <.0001 & <.0001 & <.0001 & <.0001 \\
\hline female         & -0.29657 & -0.24872 & -0.27702 & -0.20648 & -0.16802 & -0.19112 \\
        & <.0001 & <.0001 & <.0001 & <.0001 & <.0001 & <.0001 \\
\hline genderequal    & -0.20612 & -0.18739 & -0.19643 & -0.21661 & -0.18755 & -0.19917 \\
        & <.0001 & <.0001 & <.0001 & <.0001 & <.0001 & <.0001 \\
\hline age35\_54      & 0.10425 & 0.11805 & 0.09205 & 0.12067 & 0.12467 & 0.10117 \\
        & <.0001 & <.0001 & <.0001 & <.0001 & <.0001 & <.0001 \\
\hline agegt55        & 0.05357 & 0.07674 & 0.05837 & 0.11031 & 0.11144 & 0.09063 \\
        & <.0001 & <.0001 & <.0001 & <.0001 & <.0001 & <.0001 \\
\hline college        & 0.19436 & 0.16958 & 0.18600 & 0.20878 & 0.17087 & 0.18972 \\
        & <.0001 & <.0001 & <.0001 & <.0001 & <.0001 & <.0001 \\
\hline est80\_89      & -0.00630 & -0.03518 & -0.00916 & 0.00065 & -0.02310 & -0.00649 \\
        & 0.0763 & <.0001 & 0.0101 & 0.8189 & <.0001 & 0.0237 \\
\hline est90\_99      & -0.03292 & -0.07486 & -0.03773 & -0.05124 & -0.07814 & -0.06020 \\
        & <.0001 & <.0001 & <.0001 & <.0001 & <.0001 & <.0001 \\
\hline est00\_07      & -0.18201 & -0.20417 & -0.18285 & -0.18726 & -0.19766 & -0.19684 \\
        & <.0001 & <.0001 & <.0001 & <.0001 & <.0001 & <.0001 \\
\hline ecomm          & 0.01715 & 0.06257 & 0.04177 & 0.02748 & 0.06158 & 0.05666 \\
        & <.0001 & <.0001 & <.0001 & <.0001 & <.0001 & <.0001 \\
\hline agric          & 0.40638 & 0.39339 & 0.44012 & 0.11525 & 0.10500 & 0.14791 \\
        & <.0001 & <.0001 & <.0001 & <.0001 & <.0001 & <.0001 \\
\hline constr         & 0.53052 & 0.53046 & 0.53338 & 0.25166 & 0.24911 & 0.24979 \\
        & <.0001 & <.0001 & <.0001 & <.0001 & <.0001 & <.0001 \\
\hline whlsle         & 1.31265 & 1.31734 & 1.32914 & 0.29042 & 0.29642 & 0.29731 \\
        & <.0001 & <.0001 & <.0001 & <.0001 & <.0001 & <.0001 \\
\hline retail         & 0.63601 & 0.62115 & 0.64038 & -0.18361 & -0.19887 & -0.18931 \\
        & <.0001 & <.0001 & <.0001 & <.0001 & <.0001 & <.0001 \\
\hline trans          & 0.23082 & 0.22336 & 0.23228 & 0.12045 & 0.11871 & 0.11702 \\
        & <.0001 & <.0001 & <.0001 & <.0001 & <.0001 & <.0001 \\
\hline info           & 0.17076 & 0.18642 & 0.17807 & 0.26974 & 0.29783 & 0.27907 \\
        & <.0001 & <.0001 & <.0001 & <.0001 & <.0001 & <.0001 \\
\hline profsvcs       & 0.10481 & 0.11447 & 0.12067 & 0.33872 & 0.35320 & 0.35488 \\
        & <.0001 & <.0001 & <.0001 & <.0001 & <.0001 & <.0001 \\
\hline socsvcs        & -0.03031 & -0.01720 & -0.02449 & 0.20501 & 0.21724 & 0.20721 \\
        & <.0001 & 0.0007 & <.0001 & <.0001 & <.0001 & <.0001 \\
\hline persvcs        & -0.12264 & -0.13246 & -0.11912 & -0.08339 & -0.09811 & -0.08939 \\
        & <.0001 & <.0001 & <.0001 & <.0001 & <.0001 & <.0001 \\
\hline
\end{tabular}
\caption{OLS Regression with Full Set of Controls Corresponding to Table \ref{tab:4}}
\label{tab:sales_payroll_models}
\end{table}

\end{document}